\documentclass[11pt]{article}

\usepackage{acl}

\usepackage{times}
\usepackage{latexsym}

\usepackage[T1]{fontenc}
\usepackage[utf8]{inputenc}

\usepackage{microtype}

\usepackage{graphicx}
\usepackage{url}            
\usepackage{booktabs}       
\usepackage{array}          
\usepackage{amsfonts}       
\usepackage{nicefrac}       
\usepackage{xcolor}         
\usepackage{amsmath}        
\usepackage{amssymb}        
\usepackage{mathtools}      
\usepackage{amsthm}         
\usepackage{subcaption}     

\usepackage{hyperref}

\title{SALA: Semantic-Aware Logical Alignment for Complex Reasoning in In-Context Learning}

\author{
\textbf{Zhao Ji\textsuperscript{1,2}},
\textbf{Wenqing Chen\textsuperscript{1,2}}\thanks{
Corresponding author.},
\textbf{Zhixuan Chu\textsuperscript{3}},
\textbf{Jianxing Yu\textsuperscript{4,5}},
\\
\textbf{Jingping Liu\textsuperscript{1,2}},
\textbf{Shanhe Zhao\textsuperscript{6}},
\textbf{Zibin Zheng\textsuperscript{1,2}}
\\
\textsuperscript{1}School of Software Engineering, Sun Yat-sen University, Zhuhai, China \\
\textsuperscript{2}Zhuhai Key Laboratory of Trusted Large Language Models, Zhuhai, China \\
\textsuperscript{3}The State Key Laboratory of Blockchain and Data Security, \\
Zhejiang University, Hangzhou, China \\
\textsuperscript{4}School of Artificial Intelligence, Sun Yat-sen University, Zhuhai, China \\
\textsuperscript{5}Key Laboratory of Sustainable Tourism Smart Assessment Technology, \\
Ministry of Culture and Tourism, Zhuhai, China \\
\textsuperscript{6}Merchants Union Consumer Finance Company Limited, Shenzhen, China
}

\begin{document}

\maketitle

\begin{abstract}
Effective in-context learning (ICL) for complex reasoning relies on selecting the right demonstrations. Traditional retrieval methods based on surface similarity fail to capture the underlying problem-solving logic. Recent logic-based methods address this by matching predefined reasoning steps, but the rigid rules and exact-match criteria is improper to handle flexible or diverse reasoning processes.
To address the problem, we propose \textbf{SALA}, a \textbf{S}emantic-\textbf{A}ware \textbf{L}ogical \textbf{A}lignment framework. Instead of relying on a fixed inventory, SALA automatically learns task-specific reasoning operations. It then embeds these operations into a continuous semantic space and uses dynamic time warping (DTW) to align the reasoning sequences. This approach allows for soft, flexible matching of reasoning logic while remaining highly interpretable. Experiments across four reasoning benchmarks and three LLMs demonstrate that SALA outperforms existing demonstration selection methods. Further analysis confirms the roles of the operation induction and the logical semantic alignment.
\end{abstract}

\begin{figure}[!t]
\centering
\includegraphics[width=\columnwidth]{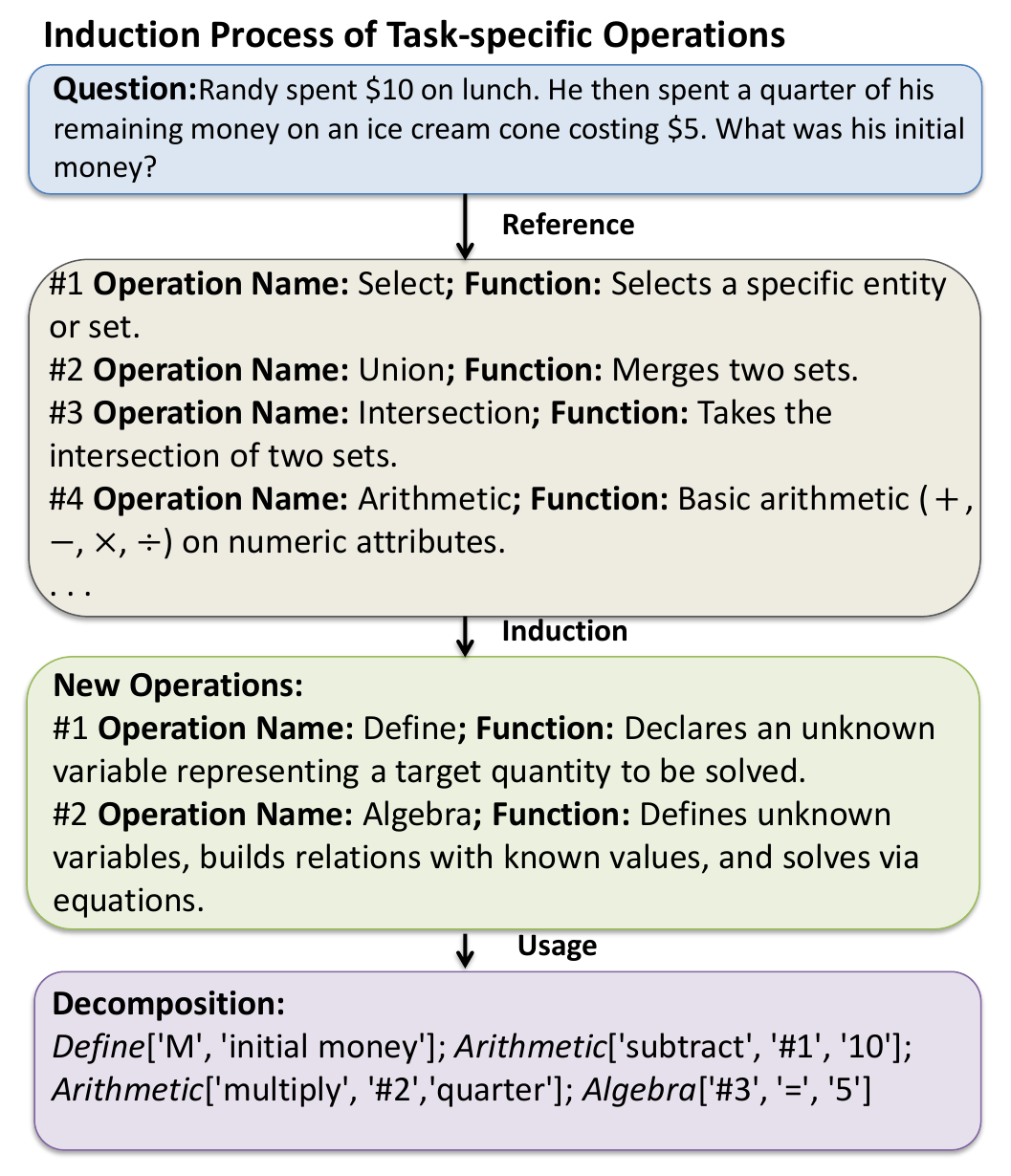}
\caption{An example of the induction process of task-specific operations.}
\label{fig:induction_example}
\end{figure}

\section{Introduction}

Large Language Models (LLMs)~\cite{10.5555/3600270.3602281,achiam2023gpt,touvron2023llama,bai2023qwen,ye2023comprehensivecapabilityanalysisgpt3,10137850} have shown strong in-context learning (ICL) ability~\cite{dong2024survey,10.5555/3666122.3667715,luo2024incontext,bhattamishra2024understanding}, enabling them to solve complex reasoning tasks from a few labeled demonstrations. 
ICL performance largely depends on demonstration selection, since useful examples can improve accuracy~\cite{ye2023compositional}, whereas mismatched or noisy examples may introduce reasoning biases and degrade performance~\cite{zhang2025learning}. The importance of selecting useful examples is further enhanced in recent agentic LLM systems, where reasoning experiences are stored as  memories and retrieved to guide future tasks~\cite{DBLP:journals/corr/abs-2509-25140}.

Existing studies have improved demonstration selection in general domain~\cite{rubin2022learning,li2023unified,ye2023compositional,yang2023representative,qin2024context}. However, most retrieval models rely on embedding similarity and the reasoning logic can be obscured by surface semantics.

The key question is how to represent demonstrations for reasoning tasks. Recent reasoning-aware methods construct intermediate representations before retrieval, including generated reasoning paths~\cite{qin2024context}, reasoning patterns~\cite{zhang2025reasoningpatterns}, latent reasoning skills~\cite{xu2024lars}, and reasoning graphs~\cite{lin2025reasoning}. These methods suggest that raw question embeddings are often insufficient, since reasoning-relevant signals can be diluted by topics, entities, and surface semantics. Problem-solving logic (PSL) takes a more symbolic route by representing each problem as a sequence of predefined reasoning operations and matching demonstrations at the operation level~\cite{ma2025problem}. Such operation-based representations reduce the influence of surface wording and provide a more explicit view of the problem-solving process.

Despite these advances, existing representations still struggle to make reasoning logic both comparable and adaptable. 
Free-form rationales or reasoning paths can express diverse reasoning processes, but their natural-language form often makes the representation noisy and unstable. 
Symbolic operation sequences, as explored by PSL, make reasoning steps easier to compare, but fixed operation spaces and exact matching rules limit their ability to generalize across task-specific and semantically equivalent reasoning patterns. 
This motivates a logical alignment framework that represents reasoning explicitly while aligning it semantically.

To this end, we propose \textbf{SALA}, a \textbf{S}emantic-\textbf{A}ware \textbf{L}ogical \textbf{A}lignment framework for reasoning-oriented demonstration selection. SALA constructs a task-adaptive operation space by inducing reasoning operations from downstream data, enabling demonstrations to be represented with problem-solving units beyond the predefined operation set. Figure~\ref{fig:induction_example} illustrates how SALA induces additional task-specific operations when the predefined operation set cannot fully express the reasoning logic of a downstream question. Then we embed operation descriptions into a continuous semantic space and applies dynamic time warping (DTW) to align reasoning-operation sequences~\cite{sakoe1978dynamic}. In this way, we keep the reasoning representation explicit while enabling soft semantic alignment beyond exact symbolic correspondence.

We conduct experiments on four reasoning benchmarks across three LLMs. 
SALA achieves better average performance against similarity-based, learning-based, and reasoning-aware demonstration selection baselines. 
Ablation studies further show that both task-adaptive operation construction and semantic sequence alignment contribute to the improvement.
The key contributions of this work are as follows:
\begin{itemize}
\item We propose SALA, a reasoning-oriented demo\-nstration selection framework that represents problem-solving logic with explicit operation sequences and aligns them in semantic space.

\item We introduce a task-adaptive operation space that extends predefined operations with reus\-able reasoning units induced from downstream data, enabling more expressive problem-sol\-ving representations without manual operation design.

\item We design a semantic DTW-based alignment strategy to measure logical similarity across reasoning-operation sequences with different lengths and decomposition granularities.

\item We validate SALA on four reasoning benchmarks and three LLMs, showing consistent average gains over recent state-of-the-art baselines and confirming the roles of operation construction and semantic alignment through ablations.
\end{itemize}

\begingroup\nolinenumbers

\begin{figure*}[t]
\centering
\includegraphics[width=0.98\textwidth]{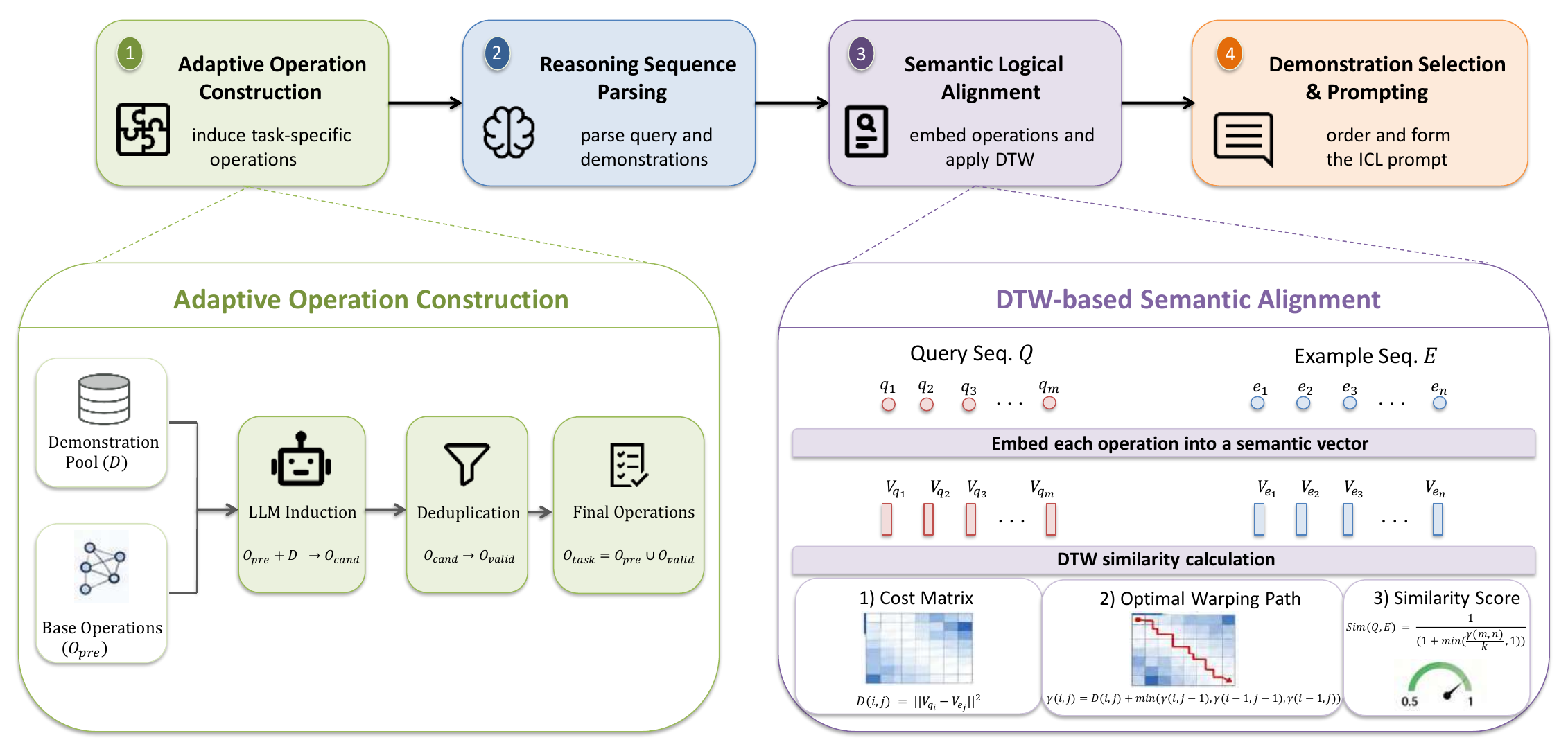}
\caption{Overview of the SALA framework. SALA constructs a task-adaptive operation set, parses queries and demonstrations into reasoning-operation sequences, and applies DTW-based semantic alignment to retrieve logically similar demonstrations for ICL prompting.}
\label{fig:overview}
\end{figure*}

\endgroup

\section{Related Work}
Existing work on ICL demonstration selec\-tion\linebreak ~\cite{dong2024survey} has studied this problem from lexical, semantic, learned, diversity-aware, and reasoning-aware perspectives. We review the most relevant lines of work in general and reasoning domains.

\subsection{General Demonstration Selection}

Early demonstration selection methods retrieve examples according to lexical overlap or sentence-level semantic similarity. Sparse retrieval methods such as BM25~\cite{10.1561/1500000019} es\-timate query-example word overlap, while embed\-ding-based methods use pretrained encoders such as BERT~\cite{devlin2019bert} to retrieve semantically similar demonstrations. Later studies improve this basic retrieval paradigm by considering supportiveness, diversity, and compositionality. For example, support-example selection chooses demonstrations according to their usefulness for the target query~\cite{li2023finding}. DPP-based selection balances relevance and diversity~\cite{yang2023representative}, and compositional exemplar selection models interactions among demonstrations beyond nearest-neighbor retrieval~\cite{ye2023compositional}. Iterative selection further shows that demonstration retrieval can be treated as a multi-step process rather than a one-shot nearest-neighbor search~\cite{qin2024context}.

Another line of work formulates demonstration selection as a learnable retrieval or optimization problem. EPR trains a retriever from language-model feedback~\cite{rubin2022learning}, while UDR learns a unified retriever for cross-task demonstration selection~\cite{li2023unified}. More recent methods use model preference, gradient matching, reinforcement learning, or many-shot optimization to improve selection quality~\cite{zhang2025learning,zhang2025selecting,wang2025demonstration,purohit2025sample}. These methods improve the retrieval of useful demonstrations, but their selection signals are still mostly derived from input similarity or model-level feedback rather than the reasoning process itself. For complex reasoning tasks,  a helpful demonstration should not only be topically or semantically related to the query, but also follow a compatible problem-solving logic.

\subsection{Reasoning Demonstration Selection}
Recent work has started to construct reasoning-oriented representations. One direction uses natural-language reasoning descriptions. Skill-KNN rewrites inputs into skill-based descriptions before applying embedding-based retrieval~\cite{an2023skill}. Luo et al.~\shortcite{luo2023dricl} extend retrieval-based ICL to chain-of-thought (CoT) prompting, and iterative demonstration selection uses generated reasoning paths to guide example retrieval~\cite{qin2024context}. These methods keep the representation flexible, but the reasoning signal is still expressed in free-form text, which can make the retrieval noisy or unstable.

A second direction introduces more structured reasoning representations. Reasoning-pattern methods select demonstrations according to task-specific reasoning patterns~\cite{zhang2025reasoningpatterns}, LaRS learns latent reasoning skills from CoT ratio\-nales\linebreak ~\cite{xu2024lars}, and RGER represents intermediate reasoning steps as reasoning graphs for exemplar retrieval~\cite{lin2025reasoning}. For mathematical reasoning, LMS3 further shows that useful demonstrations should balance semantic similarity and inference stability~\cite{liu2025what}. These methods provide stronger reasoning-oriented signals than raw question embeddings, but latent skills are less directly interpretable, and graph-based representations require more complex structure matching across examples.

PSL-guided ICL is the most closely related symbolic approach~\cite{ma2025problem}. 
It represents each problem as a sequence of predefined QDMR-style reasoning operations~\cite{wolfson-etal-2020-break} and selects demonstrations through operation-level exact matching. 
While this shows the value of explicit operation sequences for representing problem-solving logic, it still relies on a fixed operation space and rigid symbolic matching. SALA moves beyond symbolic matching by combining operation-level reasoning representations with semantic alignment. 
It constructs a task-adaptive operation space from downstream data and aligns operation sequences in semantic space with DTW, allowing explicit reasoning representations to be compared flexibly across diverse reasoning processes.

\section{Methodology}
As shown in Figure~\ref{fig:overview}, SALA follows a four-stage pipeline for reasoning-oriented demonstration selection, with adaptive operation construction and DTW-based semantic alignment serving as the two key mechanisms.
\subsection{Problem Formulation}
\label{sec:problem-formulation}

Let $\mathcal{D}=\{(q_i,y_i)\}_{i=1}^{N}$ denote the demonstration pool derived from the training set of a downstream task, where $q_i$ and $y_i$ are the question and its answer, respectively. 
Given a test query $q^\ast$, the goal is to select $k$ demonstrations from $\mathcal{D}$ that support the target LLM in solving $q^\ast$.

For reasoning tasks, we model the problem-solving logic of a question as an ordered sequence of reasoning operations. 
Given an operation set $\mathcal{O}$, a parser maps a question $q$ into
\begin{equation}
    \pi(q;\mathcal{O})=[o_1,\ldots,o_m], \quad o_j\in \mathcal{O}.
\end{equation}
After constructing the task-adaptive operation set $\mathcal{O}_{task}$, we denote the query sequence and the candidate sequence as $Q^\ast=\pi(q^\ast;\mathcal{O}_{task})$ and $E_i=\pi(q_i;\mathcal{O}_{task})$, respectively. 
Demonstration selection is then formulated as:
\begin{equation}
    \mathcal{D}_k(q^\ast)=
    \operatorname{TopK}_{(q_i,y_i)\in\mathcal{D}} Sim(Q^\ast,E_i),
\end{equation}
where $Sim(Q^\ast,E_i)$ measures the alignment between the problem-solving logic of the query and that of the candidate demonstration. 
SALA focuses on constructing the task-adaptive operation set $\mathcal{O}_{task}$ and defining $Sim(\cdot,\cdot)$ through semantic DTW alignment.

\subsection{Adaptive Construction of Reasoning Operation Sets}
To address the limited transferability of fixed predefined reasoning operation sets, we construct a task-adaptive reasoning operation set $\mathcal{O}_{task}$ by extending the 13 predefined QDMR reasoning operations $\mathcal{O}_{pre}$~\cite{wolfson-etal-2020-break} (see Appendix~\ref{appendix:qdmr-ops} for the full list) with operations induced from downstream training data. 
The construction process consists of candidate operation induction followed by two-stage deduplication.

\subsubsection{Candidate Operation Induction}
For each training question $q_i$ in the candidate demonstration pool $\mathcal{D}$, we condition an LLM on the predefined QDMR operation set $\mathcal{O}_{pre}$ and prompt it to identify additional operations only when the existing set is insufficient to cover the problem-solving logic of $q_i$. 
If no extension is needed, the model returns an empty result; otherwise, it outputs one or more candidate operation descriptions in the predefined format. 
Collecting the non-empty outputs over all questions yields a candidate operation pool $\mathcal{O}_{cand}=\{o_1,o_2,\ldots,o_M\}$, where $M$ is the number of candidate new operations. 
The full prompt used for this step is provided in Appendix~\ref{appendix:prompt-induction}.
\subsubsection{Two-Stage Deduplication of New Operations}
To avoid redundancy and ensure that newly induced operations truly complement the predefined ones, we apply a two-stage deduplication procedure.

\textbf{Stage 1: Heuristic Name-Based Deduplication}. We first perform a lightweight heuristic deduplication over the names of candidate operations. 
For each candidate operation $o_k \in \mathcal{O}_{cand}$, we extract its operation name, denoted as $n_k$. 
We then normalize the name by lowercasing it and removing spaces and underscore characters, yielding a normalized form $\tilde{n}_k$. 
Candidate entries whose extracted names contain obvious formatting artifacts or malformed markers are discarded at this stage. 
Let $\mathcal{O}_{rep}$ denote the retained representative operation set, initialized as an empty set. 
We process the candidates sequentially and compare the normalized name of the current candidate with the normalized operation names already stored in $\mathcal{O}_{rep}$.

For a current candidate $o_k$ with normalized name $\tilde{n}_k$, and an existing representative operation $\bar{o}_i \in \mathcal{O}_{rep}$ with normalized name $\tilde{n}_{\bar{o}_i}$, we apply the following deterministic rules:
\begin{itemize}
    \item If $\tilde{n}_k \subseteq \tilde{n}_{\bar{o}_i}$, we replace $\bar{o}_i$ with $o_k$.
    \item If $\tilde{n}_{\bar{o}_i} \subseteq \tilde{n}_k$, we discard $o_k$.
    \item Otherwise, we keep both entries unchanged.
\end{itemize}
If no containment relation is found between $\tilde{n}_k$ and any existing representative in $\mathcal{O}_{rep}$, we append $o_k$ to $\mathcal{O}_{rep}$. 
After all candidates are processed, $\mathcal{O}_{rep}=\{\bar{o}_1,\bar{o}_2,\ldots,\bar{o}_P\}$ becomes the representative operation set produced by Stage 1.
This heuristic removes near-duplicate operation names caused by minor formatting differences or partially overlapping name variants, while preserving the original full operation descriptions for subsequent processing. 
Because the procedure depends only on normalized names, containment checks, and candidate order, it is reproducible given the same candidate operation pool.

\textbf{Stage 2: Function Duplication Detection Based on LLM Judgment}. 
When $\mathcal{O}_{rep}=\{\bar{o}_1,\bar{o}_2,\ldots,\bar{o}_P\}$ is obtained after Stage 1, we further remove functional redundancy through a sequential judgment process. 
We initialize the task-adaptive operation set as $\mathcal{O}_{task}^{(0)}=\mathcal{O}_{pre}$. 
For each representative operation $\bar{o}_i\in \mathcal{O}_{rep}$, we construct a function comparison prompt that includes the core function description of $\bar{o}_i$ and the core function descriptions of all operations in the current operation set $\mathcal{O}_{task}^{(i-1)}$. 
The LLM outputs a binary judgment result $J(\bar{o}_i,\mathcal{O}_{task}^{(i-1)})\in\{0,1\}$, where $J(\bar{o}_i,\mathcal{O}_{task}^{(i-1)})=1$ means that $\bar{o}_i$ functionally overlaps with at least one operation in $\mathcal{O}_{task}^{(i-1)}$, and $J(\bar{o}_i,\mathcal{O}_{task}^{(i-1)})=0$ means no overlap.

The operation set is then updated recursively:
\begin{equation}
\mathcal{O}_{task}^{(i)} =
\begin{cases}
\mathcal{O}_{task}^{(i-1)} \cup \{\bar{o}_i\}, & \text{if } J(\bar{o}_i,\mathcal{O}_{task}^{(i-1)})=0,\\
\mathcal{O}_{task}^{(i-1)}, & \text{otherwise}.
\end{cases}
\end{equation}
After all representative operations are examined, the final task-adaptive reasoning operation set is $\mathcal{O}_{task}=\mathcal{O}_{task}^{(P)}$.

This sequential process ensures that each accepted operation complements both the predefined operations and the previously accepted induced operations. 
The prompts for candidate induction (Section~\ref{appendix:prompt-induction}) and deduplication (Section~\ref{appendix:prompt-dedup}) are fully documented in Appendix~\ref{appendix:prompts}.

\subsection{Construction of the Reasoning Operation Embedding Library}
To capture the functional semantic associations of operations and support semantic alignment, we convert discrete operations into continuous semantic embeddings and construct a reasoning operation embedding library, providing the foundation for subsequent similarity calculations.

First, we generate semantic description texts for each operation $o\in \mathcal{O}_{task}$, denoted as $desc(o)$. In our implementation, each $desc(o)$ describes the core function of the corresponding operation, and the description texts are used as inputs to a pretrained embedding model.

Next, we use a pretrained text embedding model as the semantic encoder, denoted as $f_{emb}(\cdot): str\rightarrow \mathbb{R}^d$, where $d$ is the embedding dimension. Specifically, the implementation uses a Hugging Face embedding model to encode the operation description text and obtains the initial semantic embedding vector through mean pooling over the hidden states:
\begin{equation}
    u_o^{init} = f_{emb}(desc(o)) \in \mathbb{R}^d
\end{equation}
We apply $\ell_2$ normalization to the initial vector before DTW matching:
\begin{equation}
    v_o = \frac{u_o^{init}}{\|u_o^{init}\|_2}
\end{equation}
where $v_o\in \mathbb{R}^d$ and $\|v_o\|_2=1$ is the final semantic embedding vector of operation $o$.

Finally, the mapping relationship, $\{(o,v_o) \mid o\in \mathcal{O}_{task}\}$, is used to form the reasoning operation embedding library $L_{emb}$. In practice, the embeddings are precomputed once for all operations in $\mathcal{O}_{task}$ and then reused during retrieval.

\subsection{DTW-Based Semantic Alignment}
In this section, we propose a semantic alignment strategy that maps reasoning operation sequences to semantic embedding sequences and uses DTW to calculate sequence-level similarity~\cite{sakoe1978dynamic}, enabling soft logical alignment across sequences of different lengths.

\subsubsection{Reasoning Operation Sequence Parsing}
Based on the constructed task-adaptive reasoning operation set $\mathcal{O}_{task}$, we use an LLM to map each query and candidate demonstration into an ordered reasoning operation sequence. Each sequence represents the problem-solving process as a series of operations drawn from $\mathcal{O}_{task}$. The parsing prompt is provided in Appendix~\ref{appendix:prompt-psl}.

\subsubsection{Semantic Sequence Generation}
We retrieve the semantic embedding of each reasoning operation in the parsed reasoning operation sequence from the embedding library, converting discrete reasoning operation sequences into continuous semantic embedding sequences. 
For the query sequence $Q^\ast = [o^\ast_1,o^\ast_2,\ldots,o^\ast_m]$, its operation embedding sequence is $\mathbf{Q}^\ast = [v_{o^\ast_1},v_{o^\ast_2},\ldots,v_{o^\ast_m}]$, wh\-ere $v_{o^\ast_i}$ is the semantic embedding of the $i$-th query operation $o^\ast_i$. 
For a candidate demonstration sequence $E_i$, we simplify it to $E=[o^E_1,o^E_2,\ldots,o^E_n]$ when computing DTW, and its semantic sequence is $\mathbf{E}=[v_{o^E_1},v_{o^E_2},\ldots,v_{o^E_n}]$, where $v_{o^E_j}$ is the semantic embedding of the $j$-th candidate operation $o^E_j$.

\subsubsection{DTW-based Similarity Calculation}
The DTW algorithm calculates the semantic similarity between the candidate semantic sequence $\mathbf{E}$ and the query semantic sequence $\mathbf{Q}^\ast$ through three steps. 
First, we construct a pairwise distance matrix $D\in \mathbb{R}^{m\times n}$, where the element $D(i,j)$ represents the semantic distance between the $i$-th vector $v_{o^\ast_i}$ in $\mathbf{Q}^\ast$ and the $j$-th vector $v_{o^E_j}$ in $\mathbf{E}$. 
Consistent with the implementation, we use Euclidean distance to measure this cost:
\begin{equation}  
D(i,j)= \|v_{o^\ast_i} - v_{o^E_j}\|_2 .
\end{equation}
where $\|\cdot\|_2$ denotes the Euclidean norm. Since the operation embeddings are precomputed in a fixed $d$-dimensional space, this distance directly measures the semantic discrepancy between two reasoning operations.

Next, we search for the optimal warping path. The warping path $W=[w_1,w_2,\ldots,w_k]$ is a sequence of coordinate pairs $(i,j)$ that satisfies three constraints to ensure a valid alignment: the boundary constraint (starting at $(1,1)$ and ending at $(m,n)$ to cover the entire problem-solving logic), the monotonicity constraint (avoiding reverse matching to maintain consistency with the reasoning process), and the continuity constraint (ensuring continuous paths without jumps to avoid missing key reasoning steps). The optimal warping path is the one with the minimum cumulative cost among all feasible paths, where the cumulative cost $\gamma(i,j)$ from the starting point $(1,1)$ to the point $(i,j)$ is calculated recursively as:
\begin{multline}
    \gamma(i,j) = D(i,j) +  \\
    \min\{ \gamma(i-1,j), \gamma(i,j-1), \gamma(i-1,j-1) \}.
\end{multline}
with the initial condition $\gamma(1,1)=D(1,1)$.

Finally, we calculate the sequence similarity. To reduce the impact of sequence length differences, we divide the cumulative DTW distance by the length of the optimal alignment path to obtain the average alignment distance, and then transform it into a bounded similarity score:
\begin{equation}
Sim(Q^\ast,E_i) = \frac{1}{1 + \min\left(\frac{\gamma(m,n)}{k}, 1\right)} .
\end{equation}

where $k$ denotes the length of the optimal warping path. Here, $\gamma(m,n)/k$ is the average DTW alignment distance per step. Since operation embeddings are $\ell_2$-normalized unit vectors, their pa\-irwise Euclidean distance lies in $[0,2]$, and the \-$\min(\cdot,1)$ clipping bounds the effective distance to $[0,1]$. Th\-erefore, $Sim(Q^\ast,E_i)\in[0.5,1]$, and a value closer to 1 indicates higher semantic similarity and greater consistency in problem-solving logic between the demonstration and the query.

We rank all demonstrations by DTW semantic similarity in descending order and select the top-$k$ most similar ones. Then, following an easy-to-hard curriculum~\cite{ma2025problem}, we sort the selected demonstrations in ascending order of reasoning operation sequence length.
\section{Experiments}

\subsection{Experimental Setup}
\subsubsection{Datasets}
We evaluate SALA on four reasoning benchmarks covering diverse task types and difficulty levels. SVAMP~\cite{patel2021nlp} contains 1,000 arithmetic word problems. GSM8K~\cite{cobbe2021training} consists of 1,319 high-quality grade-school math problems that typically require multi-step reasoning. CommonsenseQA~\cite{talmor2019commonsenseqa} is a multiple-choice commonsense reasoning benchmark with 10,881 questions, each associated with five answer options. StrategyQA~\cite{geva2021did} is an open-domain question-answering benchmark with 2,290 questions. Detailed dataset statistics and split information are provided in Appendix~\ref{appendix:dataset-stats}.

\subsubsection{Baselines}
We compare SALA with seven representative ICL baselines, including \textbf{Random} sampling, \textbf{EPR} with a contrastive retriever~\cite{rubin2022learning}, \textbf{BM25} retrieval~\cite{10.1561/1500000019}, \textbf{TopK-BERT} retrieval~\cite{devlin2019bert}, and \textbf{DPP-BERT}, which combines BERT-based retrieval with a Determinantal Point Process to balance relevance and diversity~\cite{NEURIPS2018_dbbf603f,yang2023representative}. 
We also include two reasoning-aware baselines: \textbf{PSL}, which selects demonstrations using problem-solving logic guidance~\cite{ma2025problem}, and \textbf{LMS3}, which selects demonstrations based on semantic similarity and inference stability~\cite{liu2025what}.

\subsubsection{Implementation Details}
We evaluate SALA on three LLMs: Llama3-8\-B-Instruct \cite{grattafiori2024llama}, Qwen2.5-7B-In\-struct \cite{qwen2025qwen25technicalreport}, and DeepSeek-V\-4-Pro \cite{deepseekai2026deepseekv4} (accessed via API). For each benchmark, demonstrations are retrieved from the training set. To induce task-specific reasoning operations, we use Seed-OSS-36B-Instruct \cite{seed2025seed-oss} on the training set of each benchmark. Using a separate LLM keeps operation construction independent of the evaluated target LLMs and ensures a fair comparison under the same operation space. For semantic matching, we use bert-base-uncased \cite{devlin2019bert} to encode reasoning operations. All experimental results are averaged over five runs. The number of selected examples for all baselines is based on the settings in PSL \cite{ma2025problem} for different benchmarks. Detailed configurations can be found in Appendix~\ref{appendix:hyperparams}.

\begin{table}[!t]
\centering
\resizebox{0.98\columnwidth}{!}{
\begin{tabular}{l rrrrr}
\toprule
Method & GSM8K & SVAMP & CMSQA & StrQA & Avg \\
\midrule
\multicolumn{6}{c}{\textit{Llama3-8B-Instruct}} \\
Random    & 80.91 & 85.40 & 72.91 & 82.27 & 80.37 \\
EPR       & 79.45 & 84.33 & 69.94 & 83.99 & 79.43 \\
BM25      & 79.88 & 85.53 & 69.26 & 83.49 & 79.54 \\
TopK-BERT & 80.97 & 85.00 & 71.25 & 84.13 & 80.34 \\
DPP-BERT  & 79.65 & 84.20 & 70.81 & 82.91 & 79.39 \\
PSL       & 82.03 & 84.67 & 72.89 & 84.28 & 80.97 \\
LMS3      & \textbf{82.65} & 85.40 & 73.89 & 84.89 & 81.71 \\
SALA      & 82.59 & \textbf{87.13} & \textbf{74.07} & \textbf{87.25} & \textbf{82.76} \\
\midrule
\multicolumn{6}{c}{\textit{Qwen2.5-7B-Instruct}} \\
Random    & 89.25 & 82.73 & 71.84 & 83.29 & 81.78 \\
EPR       & 87.95 & 88.00 & 74.12 & 81.95 & 83.01 \\
BM25      & 87.08 & 86.53 & 73.84 & 81.74 & 82.28 \\
TopK-BERT & 87.26 & 85.33 & \textbf{75.18} & 82.97 & 82.69 \\
DPP-BERT  & 88.07 & 83.86 & 73.59 & 83.05 & 82.14 \\
PSL       & 89.76 & 89.33 & 73.05 & 80.20 & 83.09 \\
LMS3      & \textbf{91.52} & 83.40 & 73.23 & 81.98 & 82.53 \\
SALA      & 91.16 & \textbf{91.13} & 74.73 & \textbf{83.90} & \textbf{85.23} \\
\midrule
\multicolumn{6}{c}{\textit{DeepSeek-V4-Pro}} \\
Random    & 94.54 & 90.33 & 74.45 & 90.25 & 87.39 \\
BM25      & 93.87 & 92.73 & 76.59 & 91.59 & 88.70 \\
TopK-BERT & 96.94 & 92.20 & 77.61 & 92.37 & 89.78 \\
DPP-BERT  & 95.45 & 93.00 & 75.67 & 92.58 & 89.18 \\
PSL       & 96.48 & 91.86 & 76.71 & 93.68 & 89.68 \\
SALA      & \textbf{97.12} & \textbf{94.67} & \textbf{79.28} & \textbf{94.61} & \textbf{91.42} \\
\bottomrule
\end{tabular}
}
\caption{Main results across three LLMs. EPR and LMS3 are unavailable on DeepSeek-V4-Pro (API-only access). CMSQA denotes CommonsenseQA and StrQA denotes StrategyQA. Bold indicates the best result per model--benchmark.}
\label{tab:mainresults}
\end{table}

\begin{table}[!t]
\centering
\resizebox{0.98\columnwidth}{!}{
\begin{tabular}{l rrrrr}
\toprule
Variant & GSM8K & SVAMP & CMSQA & StrQA & Avg \\
\midrule
\multicolumn{6}{c}{\textit{Llama3-8B-Instruct}} \\
w/o DTW+OPs & 81.65 & 85.00 & 72.40 & 83.41 & 80.62 \\
w/o OPs     & 82.20 & 87.06 & 73.73 & 86.20 & 82.30 \\
w/o DTW     & 82.61 & 86.87 & 74.01 & 85.09 & 82.15 \\
Full        & \textbf{82.59} & \textbf{87.13} & \textbf{74.07} & \textbf{87.25} & \textbf{82.76} \\
\midrule
\multicolumn{6}{c}{\textit{Qwen2.5-7B-Instruct}} \\
w/o DTW+OPs & 88.63 & 86.67 & 73.55 & 80.49 & 82.34 \\
w/o OPs     & 90.54 & 87.73 & 74.22 & 82.27 & 83.69 \\
w/o DTW     & 89.20 & 90.20 & 73.02 & 81.16 & 83.40 \\
Full        & \textbf{91.16} & \textbf{91.13} & \textbf{74.73} & \textbf{83.90} & \textbf{85.23} \\
\bottomrule
\end{tabular}}
\caption{Ablation study results. ``w/o DTW'' uses exact prefix matching similar to PSL; ``w/o OPs'' uses only the 13 predefined QDMR operations. CMSQA denotes CommonsenseQA and StrQA denotes StrategyQA.}
\label{tab:ablation}
\end{table}

\subsection{Main Results}
Table~\ref{tab:mainresults} summarizes the performance of SALA and the baseline methods. On Llama3-8B-Instruct, SALA achieves the best overall average and the top results on SVAMP, CommonsenseQA, and StrategyQA, while remaining competitive on GSM8K. On Qwen2.5-7B-Instruct, SALA again yields the best average, with the strongest results on SVAMP and StrategyQA. On DeepSeek-V4-Pro~\cite{deepseekai2026deepseekv4}, EPR and LMS3 are omitted as they require model training or internal hidden states inaccessible through API calls. SALA achieves the best results across all four benchmarks with an average of 91.42\%. These results show that SALA consistently improves ICL performance across models of different scales and architectures.

\subsection{Ablation Study}
To further demonstrate the effectiveness of the SALA framework, we conduct an ablation study on its key modules.

\begin{figure}[!t]
\centering
\includegraphics[width=\columnwidth]{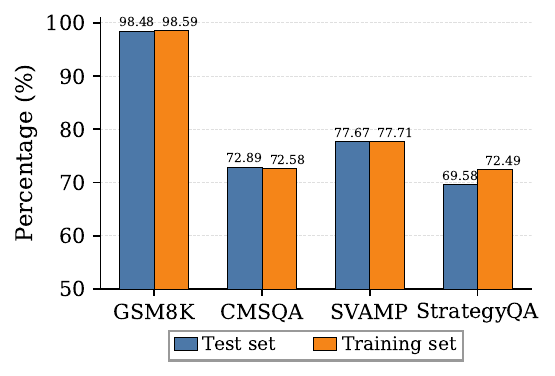}
\caption{The percentage of samples using task-specific reasoning operations in the training and test sets.}
\label{fig:task_specific_ops}
\end{figure}

\begin{figure*}[!t]
\centering
\includegraphics[width=0.88\textwidth]{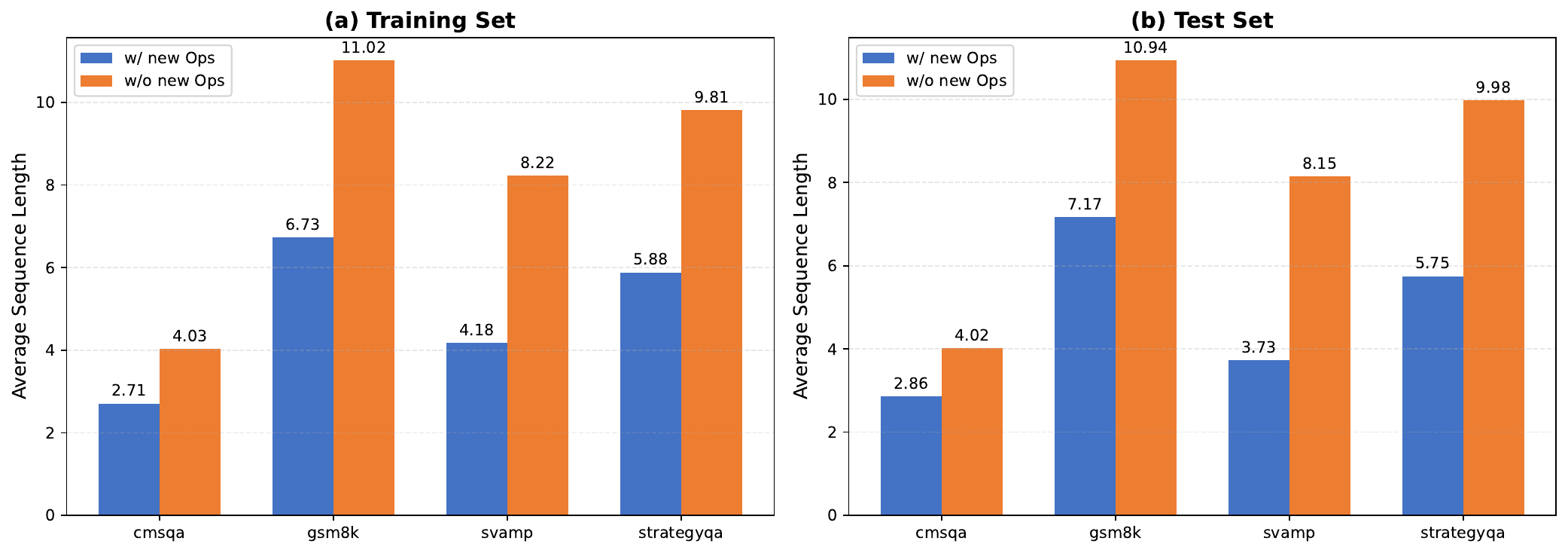}
\vspace{-3mm}
\caption{Average reasoning operation sequence length with and without task-specific operations across four benchmarks. (a) Average reasoning sequence length of training set samples; (b) Average reasoning sequence length of test set samples.}
\vspace{-4mm}
\label{fig:step_count}
\end{figure*}


As shown in Table~\ref{tab:ablation}, removing either component leads to a consistent performance drop on both models, indicating that both task-adaptive operation construction and DTW-based semantic alignment are necessary for SALA. 
When the induced operations are removed, SALA falls back to the 13 predefined QDMR operations. This reduces the average accuracy by 0.5 points on Llama3-8B-Instruct and 1.5 points on Qwen2.5-7B-Instruct, showing that the additional operations help represent task-specific reasoning patterns that are not sufficiently covered by the predefined operation set. 
When DTW is removed and exact prefix matching is used instead, the average accuracy drops by 0.6 points on Llama3-8B-Instruct and 1.8 points on Qwen2.5-7B-Instruct. This suggests that semantic sequence alignment is important for handling semantically related operations and different decomposition granularities. 
Removing both components causes the largest degradation, with average drops of 2.1 and 2.9 points on the two models, respectively. 

\subsection{SALA Analysis}
\label{sec:sala-analysis}

\subsubsection{Necessity of Inducing Task-Specific Operations}
To examine the role of task-specific reasoning operations, we measure the proportion of samples whose reasoning sequences require such operations in both the training and test sets. As shown in Figure~\ref{fig:task_specific_ops}, this proportion exceeds 68\% across all datasets. In GSM8K, it is above 98\% for both the training and test sets, whereas for the remaining datasets it generally falls between 70\% and 80\%. These findings suggest that task-specific reasoning operations are widely involved in reasoning decomposition across diverse tasks. Some representative induced operations are listed in Appendix~\ref{appendix:induced-ops}.

\subsubsection{Qualitative Case Analysis}

To further understand how each module contributes to demonstration selection, we present four case studies. Cases~1--2 isolate the effect of task-specific operations (Module~1), while Cases~3--4 isolate DTW semantic alignment (Module~2).

\paragraph{Cases~1--2: Task-Specific Operations Condense Reasoning Sequences}
Figure~\ref{fig:step_count} shows that adding task-specific operations consistently reduces average reasoning sequence length across all benchmarks. This compaction is critical for accurate demonstration matching. In Case~1 (Table~\ref{tab:case1} in Appendix \ref{appendix:case-mod1}), with Modulo, the query sequence shrinks from 9 to 7 operations, and the matched demonstration is another remainder problem; without Modulo, the inflated division-multiplication-subtraction sequence attracts a proportion problem with an incompatible reasoning pattern. In Case~2 (Table~\ref{tab:case2} in Appendix \ref{appendix:case-mod1}), with Algebra, the sequence shrinks from 13 to 7 operations; the compact Define--Algebra pattern matches another equation-solving demonstration, while the 13-step QDMR-only decomposition attracts a multi-item subtraction problem.

\paragraph{Cases~3--4: DTW Overcomes Limitations of Prefix Matching }
Cases~3--4 use only the 13 QDMR operations to isolate Module~2. Case~3 (Table~\ref{tab:case3} in Appendix \ref{appendix:case-mod2}) shows that DTW selects a percentage-markup demonstration that matches the query's reasoning pattern despite different third operations, while the longest valid prefix demonstration shares 6 of 7 operations but encodes a subtract-discount pattern incompatible with the query's add-insurance requirement. Case~4 (Table~\ref{tab:case4} in Appendix \ref{appendix:case-mod2}) shows that the longest valid prefix demonstration covers 5 of 7 operations but lacks any percentage step, causing the model to miss the tax computation entirely; DTW instead selects a semantically aligned percentage demonstration. Full case details with query examples and demonstration sequences are provided in Appendix~\ref{appendix:case-study}.




\section{Conclusion and Future Work}

We propose SALA, a reasoning-oriented demonstration selection framework that combines task-adaptive operation construction with semantic DTW-based alignment. SALA represents problem-solving logic as explicit reasoning-operation sequences, while allowing flexible matching in semantic space. Experiments on four reasoning benchmarks and three LLMs show that SALA consistently outperforms strong demonstration selection baselines. In future work, we plan to extend SALA to broader reasoning domains and explore richer reasoning structures beyond linear operation sequences.

\section*{Limitations}



SALA uses LLMs to induce task-adaptive operations and parse questions into reasoning-operation sequences. While this reduces the need for manual operation engineering, the induced operation set and parsed sequences may vary with the induction model and prompting strategy. In this work, we use a separate LLM and fixed prompts to keep the construction process consistent across experiments.

SALA currently represents problem-solving logic as a linear sequence of operations. This representation works well for the reasoning benchmarks studied in this paper, but more complex tasks may benefit from richer structures, such as hierarchical or graph-based reasoning representations. We leave the extension of SALA to these settings for future work.

\section*{Acknowledgements}
This work was supported by the National Natural Science Foundation of China (62306344, 62276279), the Guangdong Basic and Applied Basic Research Foundation (2024A1515010253, 2026A1515011800, 2024B1515020032), the Open Research Fund of the State Key Laboratory of Blockchain and Data Security, Zhejiang University (Grant No. A2537), and the Guangdong S\&T Programme Key-Area Research and Development Program of Guangdong Province (2026B0101100004).

Generative AI tools were used during the preparation and revision of this manuscript to polish the language, revise selected passages, check the consistency of terminology and reported numerical values across sections, and assist with \LaTeX{} formatting. All AI-assisted changes incorporated into the manuscript were reviewed and verified by the authors, who take full responsibility for the paper's content.
\bibliography{references_updated}

\appendix

\nolinenumbers

\section{Dataset Statistics}
\label{appendix:dataset-stats}

Table~\ref{tab:dataset_statistics} summarizes the training and test splits of the four benchmark datasets used in this paper.

\begin{table}[h]
\centering
\small
\begin{tabular}{lrr}
\toprule
Dataset & Train & Test \\
\midrule
GSM8K & 7{,}473 & 1{,}319 \\
SVAMP & 700 & 300 \\
CommonsenseQA & 9{,}741 & 1{,}140 \\
StrategyQA & 1{,}603 & 687 \\
\bottomrule
\end{tabular}
\caption{Training and test split statistics for the datasets used in our experiments.}
\label{tab:dataset_statistics}
\end{table}

\section{Predefined QDMR Reasoning Operations}
\label{appendix:qdmr-ops}

SALA extends the 13 predefined QDMR reasoning operations~\cite{wolfson-etal-2020-break} with task-specific operations. Table~\ref{tab:qdmr_operations} lists all 13 predefined operations with their core functions, example questions, and example sequences.

\begin{table*}[t]
\centering
\tiny  
\setlength{\tabcolsep}{2pt}  
\renewcommand{\arraystretch}{0.8}  
\begin{tabular}{
>{\raggedright\arraybackslash}p{1.1cm}   
>{\raggedright\arraybackslash}p{2.6cm}   
>{\raggedright\arraybackslash}p{3.0cm}   
>{\raggedright\arraybackslash}p{5.5cm}   
}
\toprule
\textbf{Operation} & \textbf{Core Function} & \textbf{Example Question} & \textbf{Example Sequence} \\
\midrule
Select & Selects a specific entity or set & How many touchdowns were scored overall? & SELECT['touchdowns']; AGGREGATE['number', '\#1'] \\
\midrule
Filter & Selects a subset matching conditions & I would like a flight from Toronto to San Diego please. & SELECT['flights']; FILTER['\#1', 'from Toronto']; FILTER['\#2', 'to San Diego'] \\
\midrule
Arithmetic & Basic arithmetic (+, $-$ , $\times$, $\div$) on numeric attributes & How many more red objects are there than blue objects? & SELECT['red objects']; SELECT['blue objects']; PROJECT['number', '\#1']; PROJECT['number', '\#2']; ARITHMETIC['difference', '\#3', '\#4'] \\
\midrule
Comparative & Selects subset greater/less than a threshold & Who are the authors with more than 500 papers? & SELECT['authors']; PROJECT['papers', '\#1']; GROUP['number', '\#2', '\#1']; COMPARATIVE['\#1', '\#3', 'more than 500'] \\
\midrule
Superlative & Selects the entity with the extreme value & What is the keyword contained by the most papers? & SELECT['papers']; PROJECT['keywords', '\#1']; GROUP['number', '\#1', '\#2']; SUPERLATIVE['\#2', '\#3', 'highest'] \\
\midrule
Aggregate & Computes mathematical properties (count, avg) of a set & How many states border Colorado? & SELECT['Colorado']; PROJECT['border states', '\#1']; AGGREGATE['number', '\#2'] \\
\midrule
Union & Merges two sets & Tell me the president and vice-president. & SELECT['president']; SELECT['vice-president']; UNION['\#1', '\#2'] \\
\midrule
Intersection & Takes the intersection of two sets & Show parties with representatives in both New York and Pennsylvania. & SELECT['representatives']; FILTER['\#1', 'in New York']; FILTER['\#1', 'in Pennsylvania']; INTERSECTION['parties', '\#2', '\#3'] \\
\midrule
Project & Obtains a specific attribute of an entity & Who is the head coach of the Los Angeles Lakers? & SELECT['Los Angeles Lakers']; PROJECT['head coach', '\#1'] \\
\midrule
Sort & Arranges elements by a specified rule & Find student addresses sorted by monthly rental. & SELECT['students']; PROJECT['addresses', '\#1']; PROJECT['monthly rental', '\#2']; SORT['\#2', '\#3'] \\
\midrule
Group & Computes properties per group element & How many female students per club? & SELECT['clubs']; FILTER['\#1', 'female students']; GROUP['number', '\#2', '\#1'] \\
\midrule
Discard & Selects subset not satisfying a condition & Find professors not playing Canoeing. & SELECT['professors']; FILTER['\#1', 'playing Canoeing']; DISCARD['\#1', '\#2'] \\
\midrule
Boolean & Determines if entity satisfies a condition & Were Scott Derrickson and Ed Wood of the same nationality? & SELECT['Scott Derrickson']; SELECT['Ed Wood']; PROJECT['nationality', '\#1']; PROJECT['nationality', '\#2']; BOOLEAN['\#3', 'the same as', '\#4'] \\
\bottomrule
\end{tabular}
\caption{The 13 predefined QDMR reasoning operations used as the base inventory in SALA.}
\label{tab:qdmr_operations}
\end{table*}

\section{Prompt Templates}
\label{appendix:prompts}

We provide the full prompt templates used in each step of the SALA pipeline. All prompts were used in English during our experiments.

\subsection{Operation Induction Prompt}
\label{appendix:prompt-induction}

This prompt instructs the LLM to induce reasoning operations not covered by the predefined 13.

\begin{quote}
\small
\textbf{Task:} Given an input question and the 13 existing reasoning operations (listed below), induce any new reasoning operations needed to solve the problem.

\textbf{Existing reasoning operations:}
\{operations\}

\textbf{Requirements:}
Supplement with new operations. Identify any new reasoning operations from the input question that are not covered by the existing ones. Describe the core functionality, provide an example question, example decomposition, and corresponding reasoning operation sequence in the same format as above. Do not output any extraneous content.
If you determine that the existing operations are sufficient to solve the input question, simply output ``No new operation needed.''

\textbf{Input question:}
\{problem\}
\end{quote}

\subsection{Sequence Analysis Prompt}
\label{appendix:prompt-psl}

This prompt decomposes a question into an operation sequence using the extended operation set.

\begin{quote}
\small
Decompose the input question into a reasoning operation sequence using the given reasoning operations.

\textbf{Given reasoning operations:}
\{operations\}

\textbf{Example}
Question: what flights are available tomorrow from denver to philadelphia?
Reasoning operation sequence: ['SELECT['flights']', 'FILTER['\#1', 'from denver']', 'FILTER['\#2', 'to philadelphia']', 'FILTER['\#3', 'if available']']
Operation name list: ['select', 'filter', 'filter', 'filter']

Follow the example format strictly and provide the reasoning operation sequence and corresponding operation name list for the input question below.

\textbf{Input question:}
\{question\}
\end{quote}

\subsection{LLM Deduplication Prompt}
\label{appendix:prompt-dedup}

This prompt checks whether a newly induced operation overlaps with existing ones (Stage 2 of deduplication).

\begin{quote}
\small
\textbf{Task:} Compare whether the functionality of the new reasoning operation overlaps with any existing reasoning operation, and determine whether the new operation is a composite operation.

\textbf{Existing reasoning operations:}
\{old\_batch\}

\textbf{Requirements:}
If the new operation's functionality overlaps with any existing operation, or if the new operation can be decomposed into existing operations (i.e., it is composite), output `<1>'. Otherwise, output `<0>'.

\textbf{New reasoning operation:}
\{new\}
\end{quote}

\subsection{ICL Prompt Template}
\label{appendix:prompt-icl}

This template formats the retrieved demonstrations and query for the final ICL inference.

\begin{quote}
\small
Before answering my question, there are some examples you can use for reference:

\textless example\textgreater
\{examples\}
\textless /example\textgreater

Here is my question:
\{question\}

Please answer it:
\end{quote}

\subsection{Evaluation Prompts}
\label{appendix:prompt-eval}

We use three dataset-specific evaluation prompts to compare candidate answers against ground-truth answers.

\textbf{Numeric Evaluation} (used for GSM8K, SV\-AMP):

\begin{quote}
\small
Your task is to compare the Candidate Answer with the Correct Answer and determine whether they are consistent.

\textbf{Question:}
\{question\}

\textbf{Candidate Answer:}
\{candidate\}

\textbf{Correct Answer:}
\{correct\}

\textbf{Criteria:}
- Numerical values are consistent if they represent the same quantity despite different formats (e.g., 0.88 is consistent with 88\%). Return <1>.
- Numerical values are also considered consistent if rounding leads to the same result (e.g., if the correct answer is 8 and the candidate is 7.96, return <1>).

\textbf{Output:} Compare whether the answers are consistent. If consistent, output <1>; otherwise, output <0>.
\end{quote}

\textbf{Text Evaluation} (used for CommonsenseQA):

\begin{quote}
\small
Your task is to compare the Candidate Answer with the Correct Answer and determine whether they are consistent.

\textbf{Question:}
\{question\}

\textbf{Candidate Answer:}
\{candidate\}

\textbf{Correct Answer:}
\{correct\}

\textbf{Criteria:}
- Evaluate whether the Candidate Answer and Correct Answer share the same meaning in the context of the given question. If they use different words but convey the same meaning, consider them consistent.

\textbf{Output:} If consistent, output <1>; otherwise, output <0>.
\end{quote}

\textbf{Boolean Evaluation} (used for StrategyQA):

\begin{quote}
\small
Your task is to compare the Candidate Answer with the Correct Answer and determine whether they are consistent.

\textbf{Question:}
\{question\}

\textbf{Candidate Answer:}
\{candidate\}

\textbf{Correct Answer:}
\{correct\}

\textbf{Criteria:}
- The Candidate Answer may be verbose while the Correct Answer is very concise. If the final conclusion of the Candidate Answer matches the boolean value (True or False) of the Correct Answer, consider them consistent.

\textbf{Output:} If consistent, output <1>; otherwise, output <0>.
\end{quote}

\section{Hyperparameter Settings}
\label{appendix:hyperparams}

Table~\ref{tab:hyperparams} lists the key hyperparameters used in our experiments.

\begin{table}[t]
\centering
\small
\setlength{\tabcolsep}{3pt}
\renewcommand{\arraystretch}{0.92}
\begin{tabular}{>{\raggedright\arraybackslash}p{2.6cm}
                >{\raggedright\arraybackslash}p{4.6cm}}
\toprule
\textbf{Parameter} & \textbf{Value} \\
\midrule
\multicolumn{2}{l}{\textit{Operation Induction}} \\
LLM for induction \& deduplication & Seed-OSS-36B-Instruct \\
Temperature & 0.01 \\
Max tokens & 4{,}096 \\
\midrule
\multicolumn{2}{l}{\textit{Semantic Embedding}} \\
Embedding model & bert-base-uncased \\
Embedding dimension & 768 \\
Max input length & 1{,}024 tokens \\
\midrule
\multicolumn{2}{l}{\textit{DTW Retrieval}} \\
Retrieval top-$k$ & 8 \\
\midrule
\multicolumn{2}{l}{\textit{Inference \& Evaluation}} \\
Target LLMs & Llama3-8B-Instruct, Qwen2.5-7B-Instruct, DeepSeek-V4-Pro \\
Temperature & 0.01 \\
Max tokens & 4{,}096 \\
Evaluation LLM & GPT-4 \\
\bottomrule
\end{tabular}
\caption{Key hyperparameter settings for SALA.}
\label{tab:hyperparams}
\end{table}

\section{Induced Reasoning Operation Examples across Downstream Tasks}
\label{appendix:induced-ops}

Table~\ref{tab:induced-ops} presents representative reasoning operations induced by SALA on each downstream benchmark through the adaptive operation set construction process.


\begin{table*}[t]
\centering
\tiny  
\setlength{\tabcolsep}{2pt}  
\renewcommand{\arraystretch}{0.8}  
\begin{tabular}{
>{\raggedright\arraybackslash}p{1.1cm}   
>{\raggedright\arraybackslash}p{2.6cm}   
>{\raggedright\arraybackslash}p{3.0cm}   
>{\raggedright\arraybackslash}p{5.5cm}   
}
\toprule
\textbf{Operation} & \textbf{Core Function} & \textbf{Example Question} & \textbf{Example Sequence} \\
\midrule
Define & Declares an unknown variable representing a target quantity to be solved & Randy spent \$10 on lunch. He then spent a quarter of his remaining money on an ice cream cone costing \$5. What was his initial money? & 
DEFINE['M', 'initial money']; ARITHMETIC['subtract', '\#1', '10']; ARITHMETIC['quarter', '\#2']; ALGEBRA['\#3', '=', '5'] \\
\midrule
Algebra & Defines unknown variables, builds relations with known values, and solves via equations & Yvonne brings a box of chocolates to school. Half have nuts and half do not. The students eat 80\% of the ones with nuts and half of the ones without nuts. If there are 28 left, how many were originally in the box? & 
DEFINE['total chocolates', '\#x']; ARITHMETIC['division', '\#x', 2]; ARITHMETIC['division', '\#x', 2]; ARITHMETIC['multiply', '\#2', '0.2']; ARITHMETIC['multiply', '\#3', '0.5']; ARITHMETIC['addition', '\#4', '\#5']; ALGEBRA['\#x', '\#6', '=', '28'] \\
\midrule
Modulo & Computes the remainder after dividing one number by another & Emma's bank account has \$100. Each day of the week she spends \$8. At the end of the week, she withdraws as many \$5 bills as possible. How many dollars remain? & 
SELECT['initial amount']; SELECT['daily spending']; SELECT['days in a week']; ARITHMETIC['product', '\#2', '\#3']; ARITHMETIC['difference', '\#1', '\#4']; MODULO['\#5', '5'] \\
\midrule
\multicolumn{4}{c}{\textbf{SVAMP}} \\
\midrule
Constant & Obtains a fixed numeric value given directly or implicitly in the problem & Mary is baking a cake. The recipe calls for 5 cups of sugar and 14 cups of flour. She already put in 11 cups of flour. How many more cups of sugar than cups of flour does she need to add now? & 
SELECT['recipe']; PROJECT['sugar', '\#1']; PROJECT['flour', '\#1']; SELECT['flour']; PROJECT['already added', '\#1']; CONSTANT['0']; ARITHMETIC['subtract', '\#2', '\#6']; ARITHMETIC['subtract', '\#3', '\#5']; ARITHMETIC['subtract', '\#7', '\#8'] \\
\midrule
Literal & Retrieves a specific numeric or attribute value directly stated in the problem & If 479 students suggested adding mashed potatoes while 489 suggested adding bacon, how many more students suggested bacon than mashed potatoes? & 
LITERAL['mashed potato', '479']; LITERAL['bacon', '489']; ARITHMETIC['difference', '\#1', '\#2'] \\
\midrule
Retain & Preserves the original attribute value of an entity when no change to that attribute is described & Ed had 12 more marbles than Doug. Ed lost 20 marbles. If Ed now has 17, how many marbles does Doug have now? & 
SELECT['Ed\'s current marbles']; SELECT['Ed\'s lost marbles']; ARITHMETIC['addition', '\#1', '\#2']; SELECT['Ed-Doug gap']; ARITHMETIC['subtract', '\#3', '\#4']; RETAIN['Doug\'s marbles', '\#5'] \\
\midrule
\multicolumn{4}{c}{\textbf{StrategyQA}} \\
\midrule
EntityLinking & Resolves a referring expression to its corresponding real-world entity & Was historical Dracula from a town in Bucharest? & 
ENTITYLINKING["historical Dracula"]; PROJECT["birthplace", "\#1"]; BOOLEAN["\#2", "in", "Bucharest"] \\
\midrule
Relate & Obtains the value of a specific relationship between two or more entities & What is the distance between Dusseldorf and Stonehenge? & 
SELECT['Dusseldorf']; SELECT['Stonehenge']; RELATE['distance', '\#1', '\#2'] \\
\midrule
Deductive & Derives a conclusion through logical reasoning from multiple factual premises & Aristotle died in 322 BC. The Model Parliament was held in 1295. The House of Lords grew out of the Model Parliament. Was Aristotle a member of the House of Lords? & 
SELECT['Aristotle']; PROJECT['death year', '\#1']; SELECT['Model Parliament']; PROJECT['held year', '\#3']; SELECT['House of Lords']; PROJECT['origin', '\#5']; BOOLEAN['\#2', 'earlier than', '\#4']; DEDUCTIVE['\#1 is a member of \#5', '\#6', '\#7'] \\
\midrule
\multicolumn{4}{c}{\textbf{CommonsenseQA}} \\
\midrule
Contrast & Obtains the opposite or contrasting concept of a given concept or attribute & The troublemaker had been hoping for a soft punishment, but the ruling handed down was quite what? & 
SELECT['soft punishment']; CONTRAST['opposite', '\#1']; SELECT['the ruling']; PROJECT['\#2', '\#3'] \\
\midrule
Similar & Finds entities or collections that are similar in type or features to a specified entity & A hurricane is similar to what other wind event? & 
SELECT['hurricane']; SIMILAR['wind events', '\#1'] \\
\midrule
Source & Locates where or how a specific attribute of an entity can be found or accessed & If I wanted to find out the hours that Marmot was open, I might look where? & 
SELECT['Marmot']; SOURCE['open hours', '\#1'] \\
\bottomrule
\end{tabular}
\caption{Representative reasoning operations induced by SALA on four downstream benchmarks.}
\label{tab:induced-ops}
\end{table*}

\section{Case Study --- Detailed Tables}
\label{appendix:case-study}

This appendix provides the full detailed tables for the four case studies summarized in Section~\ref{sec:sala-analysis} (Table~\ref{tab:case-summary}). Cases~1--2 isolate Module~1 (task-specific operations) and Cases~3--4 isolate Module~2 (DTW semantic alignment). All experiments use Llama3-8B-Instruct.

\subsection{Effect of Task-Specific Operations on Reasoning Sequence Compactness}
\label{appendix:case-mod1}

Module~1 induces task-specific operations that condense reasoning sequences which would otherwise require long chains of QDMR operations. This compactness directly affects which demonstrations are retrieved and whether the model produces the correct answer.

\subsubsection{Case 1: Modulo Condenses Remainder Computation}

\begin{table*}[t]
\centering
\small
\begin{tabular}{p{0.14\textwidth} p{0.82\textwidth}}
\toprule
\multicolumn{2}{p{0.96\textwidth}}{\textbf{Query}}\\[2pt]
\multicolumn{2}{p{0.96\textwidth}}{Emma has \$100 in her bank account. She spends \$8 each day for a full week. At the end of the week, she withdraws as many \$5 bills as possible. How many dollars remain in her account?}\\
\midrule
\textbf{Ground Truth} & \$4 \\
\textbf{w/ new OPs} & \textbf{\$4 \checkmark} \quad (with Modulo) \\
\textbf{w/o new OPs} & \textbf{\$2 $\times$} \quad (QDMR only) \\
\midrule
\multicolumn{2}{p{0.96\textwidth}}{\textbf{With New Operations (Modulo)}}\\
Query sequence (len=7) & [SELECT, PROJECT, SELECT, PROJECT, ARITHMETIC, ARITHMETIC, MODULO]\\
Matched demo & ``A factory packs 125 toys per day. Each box holds 8 toys. After filling full boxes, how many toys are left?''\\
Demo sequence (len=4) & [SELECT, PROJECT, ARITHMETIC, MODULO]\\
Analysis & This is another remainder problem; the model correctly performs the modulo computation.\\
\midrule
\multicolumn{2}{p{0.96\textwidth}}{\textbf{Without New Operations (QDMR Only)}}\\
Query sequence (len=9) & [SELECT, PROJECT, SELECT, PROJECT, ARITHMETIC, ARITHMETIC, ARITHMETIC, ARITHMETIC, ARITHMETIC]\\
& To simulate modulo, the sequence appends division $\rightarrow$ multiplication $\rightarrow$ subtraction.\\
Matched demo & ``A recipe needs 3 cups of flour per 5 servings. How many cups for 20 servings?''\\
Demo sequence (len=7) & [SELECT, PROJECT, SELECT, PROJECT, ARITHMETIC, ARITHMETIC, ARITHMETIC]\\
Analysis & This is a proportional reasoning problem, not a remainder problem. The model is misled by the similar tail of consecutive ARITHMETIC steps (5 in the query, 3 in the demo).\\
\bottomrule
\end{tabular}
\caption{Case 1: With Modulo, the compact sequence attracts another remainder problem. Without it, the inflated division-multiplication-subtraction chain attracts a proportion problem that shares the same surface pattern but has a different reasoning structure.}
\label{tab:case1}
\end{table*}

\subsubsection{Case 2: Algebra Condenses Variable Solving}

\begin{table*}[t]
\centering
\small
\begin{tabular}{p{0.14\textwidth} p{0.82\textwidth}}
\toprule
\multicolumn{2}{p{0.96\textwidth}}{\textbf{Query}}\\[2pt]
\multicolumn{2}{p{0.96\textwidth}}{Yvonne brings a box of chocolates to school. Half have nuts and half do not. The students eat 80\% of the ones with nuts and half of the ones without nuts. If there are 28 chocolates left, how many chocolates were originally in the box?}\\
\midrule
\textbf{Ground Truth} & 80 \\
\textbf{w/ new OPs} & \textbf{80 \checkmark} \quad (with Algebra) \\
\textbf{w/o new OPs} & \textbf{90 $\times$} \quad (QDMR only) \\
\midrule
\multicolumn{2}{p{0.96\textwidth}}{\textbf{With New Operations (Algebra)}}\\
Query sequence (len=7) & [DEFINE, ARITHMETIC, ARITHMETIC, ARITHMETIC, ARITHMETIC, ARITHMETIC, ALGEBRA]\\
Matched demo & ``Randy spent \$10 on lunch and a quarter of the remaining money on an ice cream cone costing \$5. What was his initial money?''\\
Demo sequence (len=5) & [DEFINE, ARITHMETIC, ARITHMETIC, ARITHMETIC, ALGEBRA]\\
Analysis & Both queries define an unknown variable and solve for it via an equation. The model recovers the algebraic structure and correctly outputs 80.\\
\midrule
\multicolumn{2}{p{0.96\textwidth}}{\textbf{Without New Operations (QDMR Only)}}\\
Query sequence (len=13) & [SELECT, SELECT, PROJECT, PROJECT, ARITHMETIC, ARITHMETIC, PROJECT, PROJECT, ARITHMETIC, ARITHMETIC, ARITHMETIC, ARITHMETIC, ARITHMETIC]\\
& Without DEFINE and ALGEBRA, each chocolate category must be separately selected and projected; remaining amounts are computed stepwise.\\
Matched demo & ``Danny has 94 guppies, 76 angelfish, 89 tiger sharks, and 58 Oscar fish. If he sells 30, 48, 17, and 24 respectively, how many fish remain?''\\
Demo sequence & [SELECT, PROJECT, SELECT, PROJECT, SELECT, PROJECT, SELECT, PROJECT, ARITHMETIC, ...]\\
Analysis & This is a multi-item subtraction problem whose SELECT--PROJECT repetition is structurally similar to the inflated query, but its reasoning logic (per-item subtraction then aggregation) is unrelated to variable solving.\\
\bottomrule
\end{tabular}
\caption{Case 2: With Algebra, the compact DEFINE--...--ALGEBRA pattern matches equation-solving demonstrations. QDMR-only decomposition produces a long chain of SELECT--PROJECT pairs that attracts structurally similar but logically unrelated multi-item problems.}
\label{tab:case2}
\end{table*}

\subsection{DTW Semantic Alignment vs.\ Prefix Subsequence Matching}
\label{appendix:case-mod2}

To isolate Module~2, both methods in this subsection use \textbf{only the 13 QDMR operations}.

\subsubsection{Case 3: Granularity Mismatch --- Best Structural Match Excluded by Prefix Constraint}

\begin{table*}[t]
\centering
\small
\begin{tabular}{p{0.14\textwidth} p{0.82\textwidth}}
\toprule
\multicolumn{2}{p{0.96\textwidth}}{\textbf{Query}}\\[2pt]
\multicolumn{2}{p{0.96\textwidth}}{Janet buys a brooch for her daughter. She pays \$500 for the material and another \$800 for the jeweler to construct it. After that, she pays 10\% of that to get it insured. How much did she pay?}\\
Query sequence (len=7) & [SELECT, PROJECT, SELECT, PROJECT, ARITHMETIC, ARITHMETIC, ARITHMETIC]\\
\midrule
\textbf{Ground Truth} & \$1{,}430 \\
\textbf{DTW} & \textbf{\$1{,}430 \checkmark} \\
\textbf{Prefix} & \textbf{\$1{,}170 $\times$} \\
\midrule
\multicolumn{2}{p{0.96\textwidth}}{\textbf{DTW-Selected Demonstration}}\\
Demo & ``John commissions a drawing. A black and white costs \$160. Color is 50\% more. How much for both?''\\
Demo sequence (len=5) & [SELECT, PROJECT, ARITHMETIC, ARITHMETIC, ARITHMETIC]\\
Reasoning & Extract base cost $\rightarrow$ apply percentage markup $\rightarrow$ sum.\\
Prefix match & \textbf{Not a valid prefix.} $c_3$ = ARITHMETIC $\neq$ $q_3$ = SELECT, so the demo's entire sequence does not match the query's prefix. Excluded.\\
\midrule
\multicolumn{2}{p{0.96\textwidth}}{\textbf{Prefix-Selected Demonstration (longest valid prefix = 6)}}\\
Demo & ``A book costs \$20 and a notebook costs \$5. With a 10\% total discount, final price?''\\
Demo sequence (len=6) & [SELECT, PROJECT, SELECT, PROJECT, ARITHMETIC, ARITHMETIC]\\
Reasoning & Sum two items $\rightarrow$ apply percentage $\rightarrow$ \textbf{subtract} (discount).\\
Prefix match & Entire demo sequence $c_{1:6}$ = $q_{1:6}$ \checkmark. Valid prefix.\\
\midrule
\multicolumn{2}{p{0.96\textwidth}}{\textbf{Analysis}}\\
& The DTW-selected demo shares the query's ``base cost $\rightarrow$ percentage'' reasoning but is excluded by prefix matching because $c_3 \neq q_3$. The prefix-selected demo matches the first 6 operations exactly, but its final reasoning step is ``subtract discount'' while the query requires ``add insurance.'' The model, misled by the subtraction pattern, outputs \$1{,}170 (\$1{,}300 $-$ \$130) instead of \$1{,}430 (\$1{,}300 + \$130).\\
\bottomrule
\end{tabular}
\caption{Case 3: DTW selects a structurally aligned demo (single-item percentage markup) despite being excluded by prefix matching due to a granularity mismatch at position 3. Prefix matching instead selects a two-item discount demo whose entire sequence matches the query's prefix, but whose final reasoning step (subtract discount) contradicts the query's requirement (add insurance).}
\label{tab:case3}
\end{table*}

\subsubsection{Case 4: Incomplete Reasoning --- Longest Valid Prefix Omits Percentage Computation}

\begin{table*}[t]
\centering
\small
\begin{tabular}{p{0.14\textwidth} p{0.82\textwidth}}
\toprule
\multicolumn{2}{p{0.96\textwidth}}{\textbf{Query}}\\[2pt]
\multicolumn{2}{p{0.96\textwidth}}{Alice bought a \$50 dress and a \$30 bag. The store applies 8\% sales tax. What is the total?}\\
Query sequence (len=7) & [SELECT, PROJECT, SELECT, PROJECT, ARITHMETIC, ARITHMETIC, ARITHMETIC]\\
\midrule
\textbf{Ground Truth} & \$86.40 \\
\textbf{DTW} & \textbf{\$86.40 \checkmark} \\
\textbf{Prefix} & \textbf{\$80 $\times$} \\
\midrule
\multicolumn{2}{p{0.96\textwidth}}{\textbf{DTW-Selected Demonstration}}\\
Demo & ``John commissions a drawing. A black and white costs \$160. Color is 50\% more. How much for both?''\\
Demo sequence (len=5) & [SELECT, PROJECT, ARITHMETIC, ARITHMETIC, ARITHMETIC]\\
Reasoning & Extract base cost $\rightarrow$ apply percentage $\rightarrow$ sum.\\
Prefix match & \textbf{Not a valid prefix.} $c_3$ = ARITHMETIC $\neq$ $q_3$ = SELECT. Excluded.\\
\midrule
\multicolumn{2}{p{0.96\textwidth}}{\textbf{Prefix-Selected Demonstration (longest valid prefix = 5)}}\\
Demo & ``A Toyota costs \$20{,}000 and a Honda costs \$25{,}000. What is the total cost?''\\
Demo sequence (len=5) & [SELECT, PROJECT, SELECT, PROJECT, ARITHMETIC]\\
Reasoning & Extract two item costs $\rightarrow$ sum them.\\
Prefix match & Entire demo sequence $c_{1:5}$ = $q_{1:5}$ \checkmark. Valid prefix.\\
\midrule
\multicolumn{2}{p{0.96\textwidth}}{\textbf{Analysis}}\\
& The DTW-selected demo teaches the model to apply a percentage after obtaining a base value, which matches the query's ``add tax'' logic. It is excluded by prefix matching because $c_3 \neq q_3$. The prefix-selected demo is a valid prefix covering 5 of 7 operations, but its reasoning stops at ``sum two costs'' with no percentage step. The model follows this incomplete pattern, outputs \$80 (the pre-tax subtotal), and misses the 8\% tax entirely.\\
\bottomrule
\end{tabular}
\caption{Case 4: DTW selects a demonstration whose reasoning pattern (base value $\rightarrow$ percentage $\rightarrow$ total) matches the query, despite not being a valid prefix. Prefix matching selects a structurally simpler demo that is a valid prefix, but whose reasoning stops at ``sum two costs'' without any percentage computation, causing the model to omit the tax step.}
\label{tab:case4}
\end{table*}

Table~\ref{tab:case-summary} summarizes the four cases.

\begingroup\nolinenumbers

\begin{table*}[t]
\centering
\small
\setlength{\tabcolsep}{3pt}  
\begin{tabular}{
  >{\raggedright\arraybackslash}p{0.06\textwidth}   
  >{\raggedright\arraybackslash}p{0.34\textwidth}   
  >{\raggedright\arraybackslash}p{0.56\textwidth}   
}
\toprule
\textbf{Case} & \textbf{What is compared} & \textbf{Key insight} \\
\midrule
1 & 13 QDMR vs.\ + Modulo & Without Modulo, remainder is simulated via 3 extra ARITHMETIC steps, attracting proportion problems instead of remainder problems. \\
\midrule
2 & 13 QDMR vs.\ + Algebra & Without Algebra, variable solving expands to 13-step SELECT--PROJECT chains, attracting multi-item subtraction problems. \\
\midrule
3 & Prefix vs.\ DTW & Prefix matching excludes the best structural match because $c_3 \neq q_3$; DTW spans the granularity gap. The longest valid prefix demo shares 6 of 7 operations but ends with ``subtract'' instead of ``add,'' misleading the model. \\
\midrule
4 & Prefix vs.\ DTW & The longest valid prefix demo matches 5 of the query's 7 operations but has no percentage step, causing the model to miss the tax computation. DTW selects a semantically aligned demo ($c_3 \neq q_3$) that includes percentage reasoning. \\
\bottomrule
\end{tabular}
\caption{Summary of the four case studies and their key insights.}
\label{tab:case-summary}
\end{table*}

\endgroup

\end{document}